\documentclass[conference,a4paper]{IEEEtran}

\usepackage{amsmath}
\usepackage{graphicx}
\usepackage{multirow}
\usepackage{cases}
\usepackage{epstopdf}
\usepackage{times}
\usepackage{color}
\usepackage{epsfig}

\def\BibTeX{{\rm B\kern-.05em{\sc i\kern-.025em b}\kern-.08em
    T\kern-.1667em\lower.7ex\hbox{E}\kern-.125emX}}
\begin{document}

\title{A New Few-shot Segmentation Network Based on Class Representation
\thanks{\textsuperscript{2*} Corresponding Author}
}

\author{\IEEEauthorblockN{Yuwei Yang\textsuperscript{1}, Fanman Meng\textsuperscript{2*},  Hongliang Li\textsuperscript{3},  King N.Ngan\textsuperscript{4}, Qingbo Wu\textsuperscript{5}\\
\textit{School of Information and Communication Engineering}\\
\textit{University of Electronic Science and Technology of China}\\
\textit{Chengdu, China}\\
\textit{\textsuperscript{1}ywyang@std.uestc.edu.cn, \{\textsuperscript{2}fmmeng, \textsuperscript{3}hlli, \textsuperscript{5}qbwu\}@uestc.edu.cn}, \textsuperscript{4}knngan@ee.cuhk.edu.hk
}
}

\maketitle

\begin{abstract}
This paper studies few-shot segmentation, which is a task of predicting foreground mask of unseen classes by a few of annotations only, aided by a set of rich annotations already existed. The existing methods mainly focus the task on ``\textit{how to transfer segmentation cues from support images (labeled images) to query images (unlabeled images)}'', and try to learn efficient and general transfer module that can be easily extended to unseen classes. However, it is proved to be a challenging task to learn the transfer module that is general to various classes. This paper solves few-shot segmentation in a new perspective of ``\textit{how to represent unseen classes by existing classes}'', and formulates few-shot segmentation as the representation process that represents unseen classes (in terms of forming the foreground prior) by existing classes precisely. Based on such idea, we propose a new class representation based few-shot segmentation framework, which firstly generates class activation map of unseen class based on the knowledge of existing classes, and then uses the map as foreground probability map to extract the foregrounds from query image. A new two-branch based few-shot segmentation network is proposed. Moreover, a new CAM generation module that extracts the CAM of unseen classes rather than the classical training classes is raised. We validate the effectiveness of our method on Pascal VOC 2012 dataset, the value FB-IoU of one-shot and five-shot arrives at 69.2\% and 70.1\% respectively, which outperforms the state-of-the-art method.

\end{abstract}

\begin{IEEEkeywords}
Few-shot Segmentation, Class Activation Map, Classification
\end{IEEEkeywords}

\section{Introduction}
Deep learning has obviously improved the performance of many computer vision tasks such as classification \cite{he2016deep}, object detection \cite{Ren2015Faster} and segmentation \cite{chen2018encoder}. However, its drawback is the serious dependence on manual annotations that are very time-consuming to be generated, especially for dense prediction tasks such as image segmentation. To this end, weakly-supervised \cite{ahn2019weakly} and semi-supervised manner \cite{lake2011one, snell2017prototypical, alfassy2019laso} attract researchers' attention. 

Few-shot segmentation is a new semi-supervised segmentation task that predicts the foreground mask of unseen classes based on few of annotations only, aided by the annotations of classes that are already existed. The key step of such task is to learn the general knowledge from known classes that can be easily extended to unseen classes. The existing methods focus on solving the problem of \textit{how to transfer segmentation cues from support images (labeled images) to query images (unlabeled images)}, and try to learn a general transformation module that has the capacity of transferring the segmentation cues from support image to query image for various classes, so that the transferred cues can be used directly to guide the segmentation of query image for unseen classes. Based on such strategy, the existing few-shot segmentation framework is build as two-branch based segmentation network, where the two branches such as support branch and query branch are used to generate features for support image and query image respectively, and transformation module is added between the two branches to transfer the segmentation cues between the two branches. Based on such framework, the existing methods try to design new transformation module that is more general and efficient, and several types of transformation modules have been proposed \cite{boots2017one, levine2018conditional}.
It is proved that the segmentation results can be enhanced by improving the transformation module. However, learning general transformation module is also proved to be a challenging task \cite{boots2017one}.

\begin{figure*}
\centering
\includegraphics[width=6.5in]{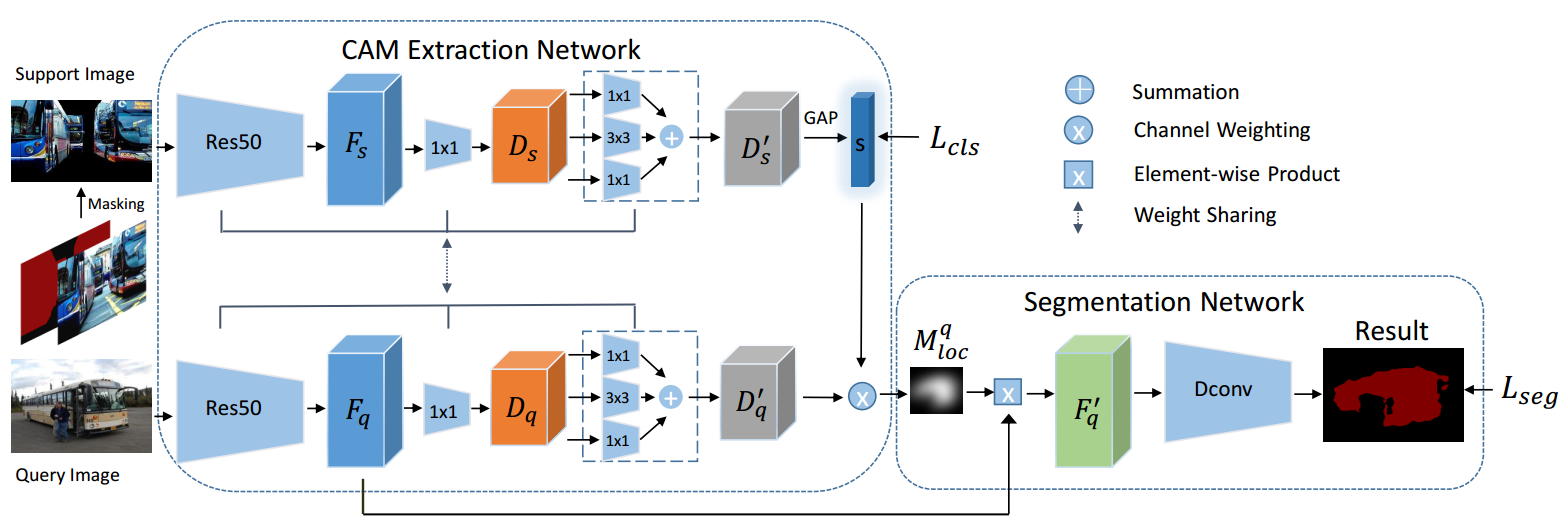}\\
 \caption{The pipeline of the proposed method. The support image and manual annotation are sent to the classification sub-network to get the classification score $S$ firstly. The query image is then sent to the classification sub-network to get class activation maps $D_q$ for each existing class. Afterwards, a refined block is adopted to get maps $D'_q$. The class activation map $M_{loc}^q$ of query image is obtained by averaging the maps $D'_q$ weighted by $S$, which is then serves as the foreground probability map to update the original query features $F_q$ to $F'_q$. Finally, the deconvolution module is used to output the segmentation result.}
\label{main_network}
\end{figure*}

Different from the existing strategy, we solve few-shot segmentation in a new perspective of ``\textit{how to represent unseen classes by existing classes}''. The idea is based on the assumption that every class can be formed by a basic attribute set $A$. By learning the basic element set $A$ from known classes, the representation of each unseen classes can also be obtained. Therefore, the prior of unseen classes can be established, and is further used to achieve the segmentation of unseen classes. In other words, unseen classes is firstly represented by existing classes with a representation module. Then, the representation is used to segment regions of unseen classes more efficiently. 

Motivated by this, this paper proposes a new few-shot segmentation network based on the strategy of representation. A new representation module in terms of class activation map is proposed. The idea is to represent the images of unseen classes based on its activation regions by the classification model of known classes. A two-branch based few-shot segmentation network is proposed. Different from the classical two branches such as support and query branch in the existing few-shot segmentation framework, the first branch is the prior generation branch that generates object prior of query image in terms of class activation map by the CAM generation module. The second branch is the segmentation branch that segments the foreground of query image based on the prior. A new CAM generation module based on the task of highlighting unseen classes rather than training classes is proposed, which firstly learns the CAM extraction module from support images of unseen classes, and then applies the CAM extraction module on query image to extract the prior map. We verify the proposed method on Pascal VOC 2012 dataset, the value FB-IoU of one-shot and five-shot arrives at 69.2\% and 70.1\% respectively, which outperforms several recent comparison methods.

\section{Proposed  Method}
\subsection{Problem Definition}
Let $P=\{(I^i_s,Y^i_s)\}^k_{i=1}$ be support images and manual annotations for unseen classes, where $k$ is the number of support images, $Y_s^i$ is the binary annotation mask for image $I^i_s$. The goal of few-shot segmentation is to build a model $f(I_q, P)$ that outputs the binary mask $\hat{M}_q$ for query images $I_q$ based on $P$, aided by a set of existing training dataset $P'=\{(I'_j,Y'_j)\}_{j=1}^{n_s}$.

\subsection{Overview}
The proposed network is shown in Fig. \ref{main_network}, which consists of two sub-networks such as the CAM generation sub-network and the segmentation sub-network. Given a support image and a query image, the first sub-network is used to generate class activation map of query image, based on the classification model of known classes. A CAM generation module is proposed. The second one outputs the segmentation mask of query image.
We next detail the two sub-networks.

\subsection{CAM Generation Module for Unseen Class}
Our goal is to represent unseen classes based on the existing classes in terms of class activation map, i.e., generating class activation map for unseen classes. Note that the classical CAM generation methods can not be used directly, as unseen classes is not considered in the classification model. Therefore, a new CAM generation module is proposed. Different from the classical CAM generation methods that use the gradient of back propagation to form the CAM, we form and learn the CAM extraction directly. We firstly learn the weight vector $S=\{s_1, s_2,...,s_n\}$ for unseen classes, where $s_i$ is the weight (similarity) of the $i$th class to unseen classes. Then, the probability map of query image is obtained by averaging the CAMs of query image highlighted by different known classes, weighted by the vector $S$. The detailed structure can be found in Fig. \ref{main_network}, where the proposed module consists of two steps such as learning the weight vector by support images, and the CAM extraction for query image based on $S$.  
\subsubsection{Learning Weights by Support Images} We intend to sufficiently use the manual annotations and support images to generate accurate CAM. A new CAM extraction module derived from the classical classification network for extracting CAM of unseen classes is proposed. The structure is shown in Fig. \ref{main_network}.  The support image is used to obtain the weight $S$. Specifically, we firstly set zero to the background pixels in order to consider the foregrounds of the manual mask only. Then, Res50 is used to extract the convolution feature of support image. Based on the last deep convolution feature of Res50, a $1\times 1$ convolution operation is applied to reduce the channel dimension of the convolution feature to the number of classes $n$, where $i$th channel means the class activation map of query image for $i$th class. We set the obtained features as $D_s$. Afterwards, a multi-scale feature extraction block is implemented to obtain final class activation map denoted as $D'_s$ that has the same size to $D_s$ . The refined block consists of feature extraction step and feature combination step. One $3\times3$ convolution operation and two $1\times1$ convolution operation are used to extract multi-scale features, and the multi-scale features are combined to obtain the refined class activation map. Finally, a global average pooling is applied on the multiple-scale class activation map to get the weight vector $S$. 

Note that the proposed CAM extraction module is based on the classification network pre-trained on known classes. Here, we use the loss function Eq.\ref{cls_loss} to supervise the learning of the classification network, i.e.,
\begin{equation}
L_{cls}=log{(1+exp(-\hat{S} \cdot S))}\label{cls_loss}
\end{equation}
where $\hat{S}$ is the class-level labels, and $S$ is the classification score.

It is seen that the weight $S$ is very important to the CAM generation for unseen class. Here, although the extraction of weight is learned automatically, it can be simply considered as the similarities between unseen class and training classes. Therefore, the CAM of unseen class can be obtained by the sum of the CAMs of training classes
weighted by their similarities.

\subsubsection{Extracting CAM for Query Images}
Given a query image of unseen class, we forward it to the classification network (Res50) to obtain the convolution feature $F_q$. A $1\times  1$ convolution layer is then applied on $F_q$ to obtain $D_q$ with channel number $n$. After that, we implement the multi-scale feature extraction block to obtain feature $D'_q=\{D_1, D_2,...,D_n\}$ with the same size to $D_q$. Then, each channel feature of $D'_q$  is weighted with the weight vector $S$, and the weighted channel features are summed to obtain the foreground prior $M_{loc}^q$ of query image. Such process can be represented by
\begin{equation}
M_{loc}^q = \sum_{i=1}^n D_i \cdot s_i
\end{equation}
The foreground prior $M_{loc}^q$ is then forwarded to the segmentation network for the foreground prediction. It is seen that the object regions of unseen class is obtained based on the relationships between unseen class and known classes.

\begin{figure}
\centering
\includegraphics[width=3in, height=0.8in]{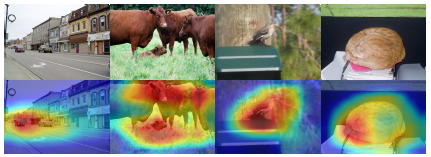}\\
 \caption{The class activation maps of images from unseen classes by the proposed method. It is seen that the proposed method can generate class activation maps of unseen classes successfully.}
\label{class activation map}
\end{figure}

Some class activation maps of unseen class can be found in Fig. \ref{class activation map}. It is seen that the object regions of unseen class are highlighted successfully, which demonstrates the effectiveness of the proposed CAM based representation module.

\subsection{The Network of Few-shot Segmentation}
After obtaining the class activation map $M_{loc}^q$, we next normalize it into [0, 1] by
\begin{equation}
\bar{M_{loc}^q} = \frac{M_{loc}^q-min(M_{loc}^q)}{max(M_{loc}^q)-min(M_{loc}^q)}
\end{equation}
Then, the normalized class activation map serves as an attention module to weight the feature $F_q$  of query image extracted from the backbone network. The filtered feature is then forwarded to a simple deconvolution block to obtain the final segmentation result $\hat{M}_q$. Such process can be represented as 
\begin{equation}
\hat{M}(q) = Dconv(F_q*  Cat(\underbrace{\bar{M_{loc}^q},..., \bar{M_{loc}^q}}_{\text{total of $n_F$ items}}))
\end{equation}
where $n_F$ is the number of channels of $F_q$. $Cat(\cdot)$ is the concatenation operation. The loss function $L_{seg}$ is set to the cross-entropy loss.

\begin{table}[h]
        \centering
        \caption{The details for classifying the 20 classes into four sub-datasets. There are 4 sub-datasets, and $PASCAL-5^i$ represents the $i$th subset, where $i=\{0, 1, 2, 3\}$.}
        \label{table_subdataset}
        \begin{tabular}{cc}
                \hline
                sub-dataset &      corresponding classes\\
                \hline
                $PASCAL-5^0$&   aeroplane, bicycle, bird, boat, bottle\\
                $PASCAL-5^1$&   bus, car, cat, chair, cow\\
                $PASCAL-5^2$&   diningtable, dog, horse, motorbike, person\\
                $PASCAL-5^3$&   potted plant, sheep, sofa, train, tv/monitor\\
                \hline
        \end{tabular}
\end{table}
\subsection{Training and Inference}
In training stage, because the proposed network is an end-to-end network, we train the network based on known classes directly. The details of the training setting can be found in Section \ref{train}. It is worth noting that the class activation map is implicitly represented by feature $D$ and $D'$ here. Therefore, the CAM extraction manner for unseen class is learned automatically and directly without back propagation. 

In the reference stage, the segmentation result is obtained directly by the network without fine-tuning.
\section{Experiments}
 We implement the proposed network on Pytorch. Adam optimizer is used to update parameters. One Nvidia Titan XP GPU is used. We set learning rate to 1e-4 which decays 0.7 times per 10 epochs. Our backbone is set to Res50 pre-trained on ImageNet, and the top three layers is frozen during training. The size of input image is $320\times 320$.
\subsection{Implementation Details}\label{train}
We implement experiment on Pascal VOC 2012 \cite{Everingham2015The} dataset and its augmentation dataset SBD \cite{hariharan2011semantic}. Similar to \cite{boots2017one}, we split the 20 classes into 4 sub-datasets, each of which contains 5 classes. Details can be found in Table \ref{table_subdataset}. For the four sub-datasets, one is selected as the unseen dataset for evaluation, the other three are used as known datasets for training. The image pairs for training are randomly selected from the training dataset. For fair comparison with the existing methods, we use the same seed for random sampling, and select the same 1000 image pairs in the testing stage. In the training stage, we use two of the three sub-datasets(ten classes) to train our classification network, and the rest one sub-dataset (five classes) as unseen classes to train the proposed network.  

\begin{figure}
\centering
\includegraphics[width=3.5in, height=1.5in]{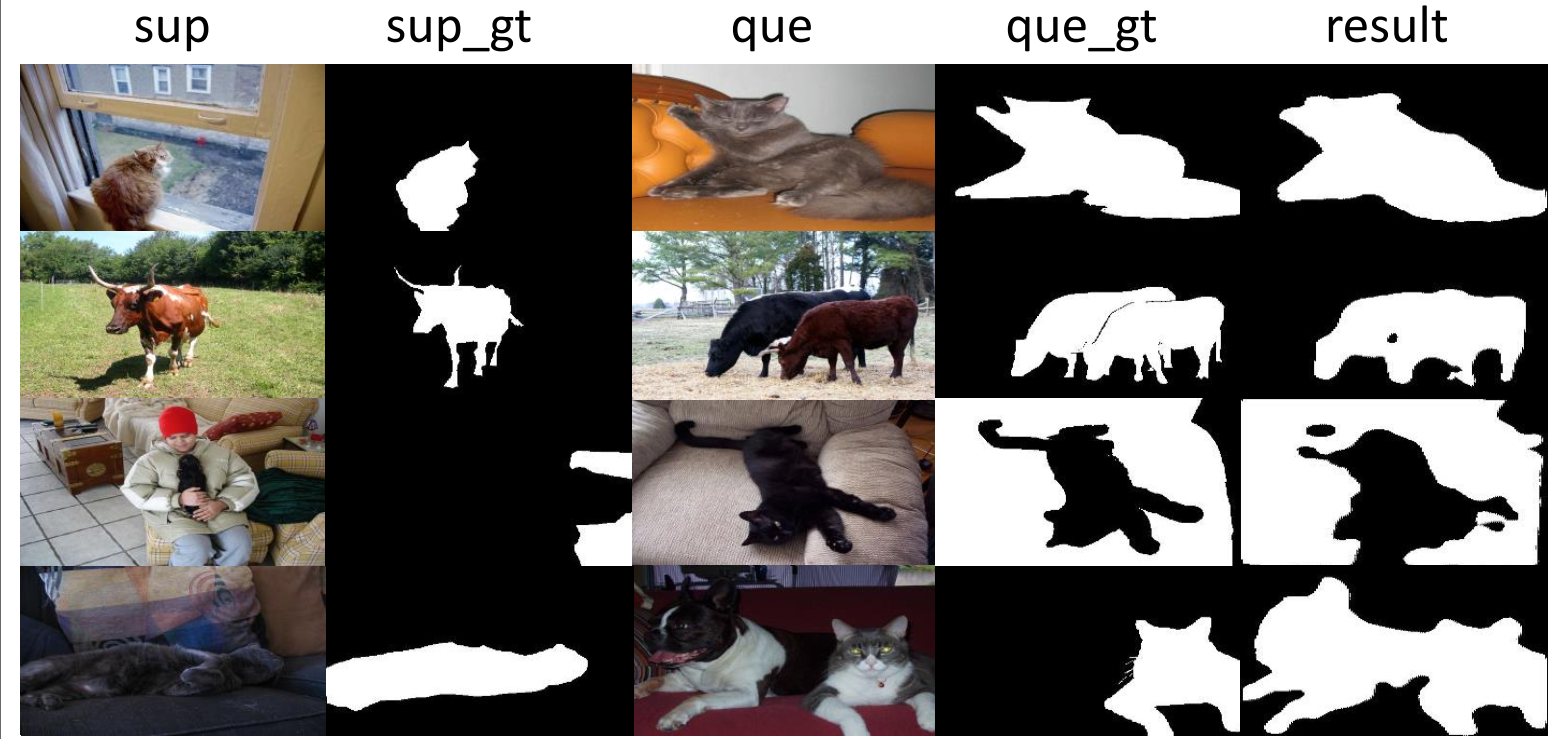}
 \caption{The subjective results of the proposed method. From left to right: support images, ground-truth of support images, query images, ground-truth of query images and segmentation results, respectively.}
 
\label{sub_reslut}
\end{figure}

\begin{table*}
\centering
        \caption{The FB-IoU values of One-shot and Five-shot segmentation by the proposed method and the comparison methods on Pascal VOC 2012 dataset. The best results are in bold.}
        \label{fb_iou}
        \begin{tabular}{ccccccccc}
                \hline
                Methods&  FG-BG\cite{levine2018conditional} & OSLSM\cite{boots2017one}&  co-FCN\cite{levine2018conditional}&  PL\cite{dong2018few}& SG-One\cite{DBLP:journals/corr/abs-1810-09091} &A-MCG\cite{aaai19:hut} &CA-Net\cite{CVPR19Zhang5}& Ours\\
                \hline
                One-shot &55.1&61.3&60.1&61.2&63.1&61.2&66.2&69.2\\
                \hline
                Five-shot &55.6&61.5&60.2&62.3&65.9&62.2&69.6&70.1\\
                \hline
        \end{tabular}
\end{table*}


The FB-IoU \cite{levine2018conditional}  that calculates mean intersection over union of both foreground and background is used for objective evaluation. 

\subsection{Subjective Results}

The subjective results of the proposed method are shown in Fig. \ref{sub_reslut}. The support images, the ground-truth of support images, the query images, the ground-truth of query images and the segmentation results are displayed from left column to right column, respectively. The first three rows show successful results. It is seen that the proposed method segments objects from these images successfully. Meanwhile, the last row displays some case of failures, where the region of ``Dog'' is wrongly segmented as ``Cat''. This is caused by the fact that ``Cat'' and ``Dog'' are very similar so that it intends to segment both of the object regions as foreground.

\subsection{Objective Results and the Comparisons with Benchmarks}
We next display the objective results in terms of FB-IoU value. In addition, we compare the proposed method with several recent few-shot segmentation methods. The results are displayed in Table \ref{fb_iou}, where One-shot and Five-shot segmentation are considered. It is seen that the FB-IoU of One-shot segmentation on the four evaluation sub-dataset is 69.2\%, which is better than the comparison methods. In addition, the value of the FB-IoU of Five-shot is 70.1\%, which also outperforms the comparison methods. This demonstrates the effectiveness of the proposed method.

\section{Conclusion}
In this paper, a new few-shot segmentation strategy based on class representation is proposed. A novel few-shot segmentation network is established. The proposed segmentation network consists of two branches. One is CAM generation network that obtains the class activation map of query image based on the classification model pre-trained on known image and support image of unseen class. The other is segmentation network that segments foreground from query image based on the class activation map. A new CAM generation module for unseen class is proposed. The proposed method is verified on Pascal VOC dataset. Experimental results demonstrate the effectiveness of our proposed method with larger FB-IoU values.

\section*{Acknowledgment}
This work was supported in part by the National Natural Science Foundation of China under Grant 61871087, Grant 61502084, Grant 61831005, and Grant 61601102, and supported in part by Sichuan Science and Technology Program under Grant 2018JY0141.

\bibliographystyle{IEEEtran}
\bibliography{egbib.bib}

\end{document}